\documentclass[lettersize,journal]{IEEEtran}
\usepackage{amsmath,amsfonts}
\usepackage{algorithmic}
\usepackage{algorithm}
\usepackage{array}
\usepackage[nolist]{acronym}
\usepackage[caption=false,font=normalsize,labelfont=sf,textfont=sf]{subfig}
\usepackage[bottom]{footmisc}
\usepackage{textcomp}
\usepackage{stfloats}
\usepackage{url}
\usepackage{verbatim}
\usepackage{graphicx}
\usepackage{xcolor}
\usepackage{cite}
\usepackage[separate-uncertainty=true,uncertainty-separator={\pm}]{siunitx}
\begin{acronym}
    \acro{NN}[NN]{Neural Network}
    \acro{CNN}[CNN]{Convolutional Neural Network}
    \acro{SNN}[SNN]{Spiking Neural Network}
    \acro{ROI}[ROI]{Region of Interest}
    \acro{EKF}[EKF]{Extended Kalman Filter}
    \acro{EM}[EM]{Expectation Maximization}
\end{acronym}

\makeatletter
\let\NAT@parse\undefined
\makeatother
\usepackage[colorlinks,citecolor=red,urlcolor=blue,bookmarks=false,hypertexnames=true]{hyperref}
\usepackage[capitalise]{cleveref}

\begin{document}

\title{An Event-Based Perception Pipeline for a Table Tennis Robot}

\author{Andreas Ziegler$^{1}$, Thomas Gossard$^{1}$, Arren Glover$^{2}$, and Andreas Zell$^{1}$
\thanks{$^{1}$Andreas Ziegler, Thomas Gossard, and Andreas Zell are with the University of Tübingen \{andreas.ziegler, thomas.gossard, andreas.zell\}@uni-tuebingen.de $^{2}$Arren Glover is with the Istituto Italiano di Tecnologia, Italy}
\thanks{This research was partially funded by Sony AI.}}



\maketitle

\begin{abstract}
Table tennis robots gained traction over the last years and have become a popular research challenge for control and perception algorithms. 
Fast and accurate ball detection is crucial for enabling a robotic arm to rally the ball back successfully. 
So far, most table tennis robots use conventional, frame-based cameras for the perception pipeline.
However, frame-based cameras suffer from motion blur if the frame rate is not high enough for fast-moving objects.
Event-based cameras, on the other hand, do not have this drawback since pixels report changes in intensity asynchronously and independently, leading to an event stream with a temporal resolution on the order of $\mathbf{\mu}$s.
To the best of our knowledge, we present the first real-time perception pipeline for a table tennis robot that uses only event-based cameras.
We show that compared to a frame-based pipeline, event-based perception pipelines have an update rate which is an order of magnitude higher.
This is beneficial for the estimation and prediction of the ball's position, velocity, and spin, resulting in lower mean errors and uncertainties.
These improvements are an advantage for the robot control, which has to be fast, given the short time a table tennis ball is flying until the robot has to hit back.

\end{abstract}


\section*{Supplementary Material}
Additional resources are available at: \url{https://cogsys-tuebingen.github.io/event-based-table-tennis}

\section{Introduction}
\IEEEPARstart{I}{n} recent years, robotic table tennis has become a popular research challenge for control and perception algorithms.
While not yet able to compete with professional players, table tennis robots are an exciting research environment to bring perception and control algorithms towards their limits~\cite{Ziegler2023corlw}\cite{Ambrosio2024arxiv}.
Fast and accurate ball detection is a crucial perception task for a table tennis robot system.
So far, most research uses frame-based cameras together with a \acf{CNN} based ball detection or a classical computer vision approach~\cite{Tebbe2019gcpr}\cite{DAmbrosio2023rss}\cite{GomezGonzalez2019robotics}\cite{Ding2022iros}.

While frame-based cameras are the de-facto standard visual perception sensor, they do suffer from motion blur if the camera's frame rate is not high enough for fast-moving objects.
One way to counteract this limitation is by using cameras with a higher frame rate.
However, a higher frame rate leads to more overall data that needs to be processed, increasing the computational resources required.
Therefore, the bottleneck shifts from the sensor to the algorithm.
Another way to approach this challenge is to make use of event-based cameras, also known as dynamic vision sensors (DVS)~\cite{Gallego2020pami}.
\begin{figure}[t!]
    \centering
    \includegraphics[width=1.0\linewidth]{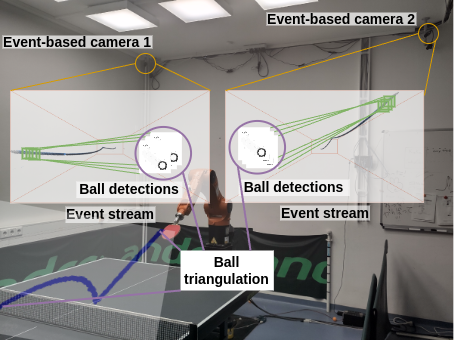}
    \caption{
        Background: The industrial robot arm of our table tennis robot setup for which the proposed perception pipeline is designed.
        The two event-based cameras, indicated with {\color{orange}orange circles}, are mounted on the ceiling.
        Foreground: The event streams of the two event-based cameras with detected balls on the EROS event surface, visualized {\color{green}in green}, the triangulation process (using the ball detections from both cameras and calculating the 3D position given the camera calibration), indicated {\color{violet}in violet}, and the resulting triangulated 3D trajectory, shown {\color{blue}in blue}.
    }
    \label{fig:eye-catcher}
\end{figure}
Event-based cameras report logarithmic changes in intensity if the brightness change exceeds a specified threshold, asynchronously and for every pixel independently, with so-called \textit{events}.
The asynchronous event-based nature of event-based cameras enables a high temporal resolution, in the order of $\mu s$, a low latency, and a high dynamic range, making them particularly suitable for capturing fast-moving objects~\cite{Monforte2020aicas}\cite{Mitrokhin2018iros}\cite{Forrai2023icra} and scenes with a high dynamic range~\cite{Perot2020neurips}\cite{Stoffregen2020eccv}.
Events are tuples of the form $<$$x, y, t, p$$>$ with the location of the event $(x, y)$, the time stamp $t$ and the polarity of the event $p \in \{-1, 1\}$, indicating if the pixel got darker or brighter.
Since this data format is fundamentally different from the frames of frame-based cameras, new algorithms are required to make use of the event-based camera's advantages.

To the best of our knowledge, we present the first real-time perception pipeline for a table tennis robot running entirely with event-based cameras and using their event stream as input.
Thus, we address the fundamental task of ball detection in the perception pipeline of a table tennis robot but leave out the robot control part since it is out of scope for this work.

In event-based computer vision, there is a trade-off between processing single events and batches of events.
While a single event does not contain enough information to be useful, processing it is usually fast.
On the other hand, event batches contain enough information, but batching introduces a latency, and processing batches tends to be more time-consuming compared to single events.
Additionally, using event batches introduces the need to adjust the size of the batch based on the velocity of the perceived objects.
Fast moving objects will generate more events than slow moving objects in the same amount of time.
If the velocity of objects varies, batching does not create consistent representations.

We maintain a low latency while increasing observed information by processing the incoming events in one thread on an \textit{event-by-event} basis and perform the more compute intensive perception task in a second thread \textit{as-fast-as-possible}, following previous work~\cite{Glover2022pami}\cite{Glover2024icra}.
This allows the algorithm to make use of event history and, therefore, preserve all the information contained in the event stream.

An event representation that supports such an \textit{event-by-event} update, without the need to recompute the whole representation, is the EROS event representation, used in~\cite{Glover2024icra}.
We use EROS in combination with a fast circle detector for the ball detection in our perception pipeline.

\textbf{Contributions} of this work are as follows:
\begin{itemize}
    \item We present the first real-time event-based perception pipeline for a table tennis robot
    \item With our fast ball detector, our event-based perception pipeline achieves an order of magnitude more position estimates of the ball compared to frame-based pipelines
    \item We show that this increased position update rate leads to lower uncertainties of the estimated positions, velocities, and spin of the ball's trajectory, which is beneficial for the robot control
    \item We make the developed perception pipeline publicly accessible on our project page
\end{itemize}

\section{Related Work}

In this related work, we will first give an overview of the latest research around table tennis robots in~\cref{subsec:rl_ttr}.
Next, we cover event-based object detection in~\cref{subsec:rel_ebod} and event representations in~\cref{subsec:rel_er}.

\subsection{Table Tennis Robots}\label{subsec:rl_ttr}

Ever since Billingsley initiated a robot table tennis competition in 1983~\cite{Billingsley1983robot}, robotic table tennis has been a popular tool for research in computer vision and robot control.
A table tennis robot using frame-based cameras with a \ac{CNN}-based ball detection was presented in~\cite{GomezGonzalez2019robotics}.
In~\cite{Buchler2022tor}, the authors have designed a completely new pneumatic robot arm able to attain very high-end-effector speeds.
The authors in~\cite{Ji2021iros} introduced a table tennis robot system but focused on trajectory prediction and hitting velocity control.
Unfortunately, they do not explain their vision system in detail.

In~\cite{DAmbrosio2023rss}, Google DeepMind presented their table tennis robot system, including perception, planning, and robot control.
In a recent update~\cite{Ambrosio2024arxiv}, the same authors have shown that while not yet able to compete with professional players, table tennis robots can compete with amateurs and intermediate-level players.
Table tennis robots are also widely used as a use case for reinforcement learning, e.g., in~\cite{DAmbrosio2023rss}\cite{Ding2022iros}\cite{Yang2021ieee}\cite{Gao2020iros}.

All of the above table tennis robot systems use frame-based cameras for their perception system and do not make use of event-based cameras.
In this work, we developed a perception pipeline for a system similar to~\cite{Tebbe2019gcpr}, using only event-based cameras.
We show the advantages of event-based perception pipelines compared to a perception pipeline using conventional, frame-based cameras.

After the ball detection, the next step in a table tennis robot system is the prediction of the ball's trajectory.
Given the observation of the positions of the ball until the current time, the robot system needs to predict the ball's future trajectory to plan the hitting stroke of the robot.
Even for humans, this prediction is difficult and requires years to get a good estimation of balls with heavy spin.
This is in particular due to the difficulty of measuring spin~\cite{Tebbe2019gcpr}.

Next to the more traditional approaches of curve fitting and an \ac{EKF} in~\cite{Tebbe2019gcpr}, a GRU-based learning approach was used in~\cite{Gao2022ijcnn}.
In~\cite{Achterhold2023l4dc}, the authors use an \ac{EKF} and learn the parameters of the dynamics model outperforming two black-box approaches.
All these works state the importance of accurate spin and a low spin uncertainty.

%
%
%

\subsection{Event-Based Object Detection}\label{subsec:rel_ebod}

Event-based object detection can be subdivided into model-based and learning-based methods.
Most learning-based methods make use of deep learning architectures to learn object detection tasks given a training dataset.
In~\cite{Perot2020neurips}, Prophesee presented the first event-based object detector that does not use an intermediate gray scale representation.
%
%
Recurrent layers are used to take the temporal information of the event data into account.
An event-based convolutional layer (e-conv) and an event-based max pooling layer (e-max-pool) were introduced in~\cite{Cannici2019cvprw}.
Inference time was accelerated using a sparse update method in which only hidden and output neurons changed by incoming events were processed.
To make better use of the data structure of the events stream, Graph Neural Networks (GNNs) were used in~\cite{Schaefer2022cvpr}.
The incoming events iteratively update the graph rather than fully reconstructing it at each time-step, preserving the asynchronous nature of event-based cameras.
Also, more recent network architectures such as Vision Transformers (ViTs) were introduced for event data~\cite{Liu2024wacv}.
While learning-based methods tend to achieve higher accuracy, their inference time is often longer than the runtime of model-based methods.

An early model-based approach for object detection made use of the Hough transform to track objects with circular structure~\cite{Glover2016iros}.
In later work, the same authors implemented a particle filter with an observation heuristic, which can be adjusted depending on the object to track~\cite{Glover2017iros}.
In~\cite{Forrai2023icra} and~\cite{Falanga2020sr}, events are clustered using DBSCAN to differentiate between the background and independently moving objects with a motion-compensated mean time stamp image.
In our setup, the cameras are static, and therefore, motion compensation, which tends to be computationally heavy, is not needed.

Since a fast runtime is essential in our table tennis robot setup, we use a model-based approach with an event representation that allows \textit{event-by-event} updates.
Event-based cameras produce a lower-latency ball trajectory prediction as information is available in the system earlier than traditional cameras~\cite{Monforte2020aicas}.

\subsection{Event Representation}\label{subsec:rel_er}

Depending on the algorithm, events are either consumed \textit{event-by-event} or in a \textit{batch-of-events}.
The minority of work does \textit{event-by-event} processing, mostly filter-based approaches, like~\cite{Scheerlinck2019accv}, and \acp{SNN}, like~\cite{Seifozzakerini2016bmcv}.
If events are processed in batches, there is a variety of choices on how to represent the information present in the event data.
For the sake of completeness, we cover the relevant ones here, where we take the definition from~\cite{Gallego2020pami}.

The simplest form is event packets, where events in a spatio-temporal neighborhood are processed together.
In event packets, the precise time stamp and polarity information is retained.
The next, almost obvious choice is the 2D histogram or event frame.
The events in a time window are converted in a simple way into an image/frame that can be fed to image-based computer vision algorithms.
The temporal information present in event data is used in the so-called time surface (TS).
In a time surface, a 2D map, each pixel stores a single time value.
Therefore, the ``intensity'' of the converted image is a function of the motion history at that location, with larger values corresponding to a more recent motion.
The concept of TS was further improved in~\cite{Lagorce2017pami} by introducing a hierarchical representation.
In~\cite{Sironi2018cvpr}, the authors present Histograms of Averaged Time Surfaces (HATS), which is less sensitive to noise and non-idealities of event-based cameras.

Instead of using hand-crafted event representations, the authors in~\cite{Gehrig2019iccv} introduced a combination of filters and convolutions, which can be learned in an end-to-end fashion when used in \acp{NN}.

In~\cite{Manderscheid2019cvpr}, the authors use a speed invariant time surface for learning to detect corner points with a simple random forest, showing the benefits of a speed invariant representation.
A more recent event representation is the threshold-ordinal surface (TOS), introduced in~\cite{Glover2022pami}.
TOS is designed to accumulate visual data, so it is compatible with conventional image processing algorithms like, e.g., circle or corner detection.
The representation provides a coherent and bound spatial representation of the asynchronous events, partially maintaining the information about their temporal order and attempting to capture the most up-to-date position of edges in the scene.
In this work, we make use of an improved version of TOS, the Exponential Reduced Ordinal Surface (EROS), used in~\cite{Glover2024icra}.
EROS can be used at any given point in time by downstream processing, with values between $0$ and $255$.

\section{Methodology}\label{sec:methodology}

In this part, we introduce our event-based perception pipeline.
We start with a description of the setup in~\cref{subsec:setup}.
Then, we give an overview of the perception pipeline in~\cref{subsec:perception-pipeline}.
In~\cref{subsec:event-update}, we explain the \textit{event-by-event} update of our approach.
In the last part, the \textit{as-fast-as-as-possible} ball detection is covered in~\cref{subsec:ball-detection}.

\subsection{Setup}\label{subsec:setup}

\begin{figure}[t!]
    \centering
    \includegraphics[width=1.0\linewidth]{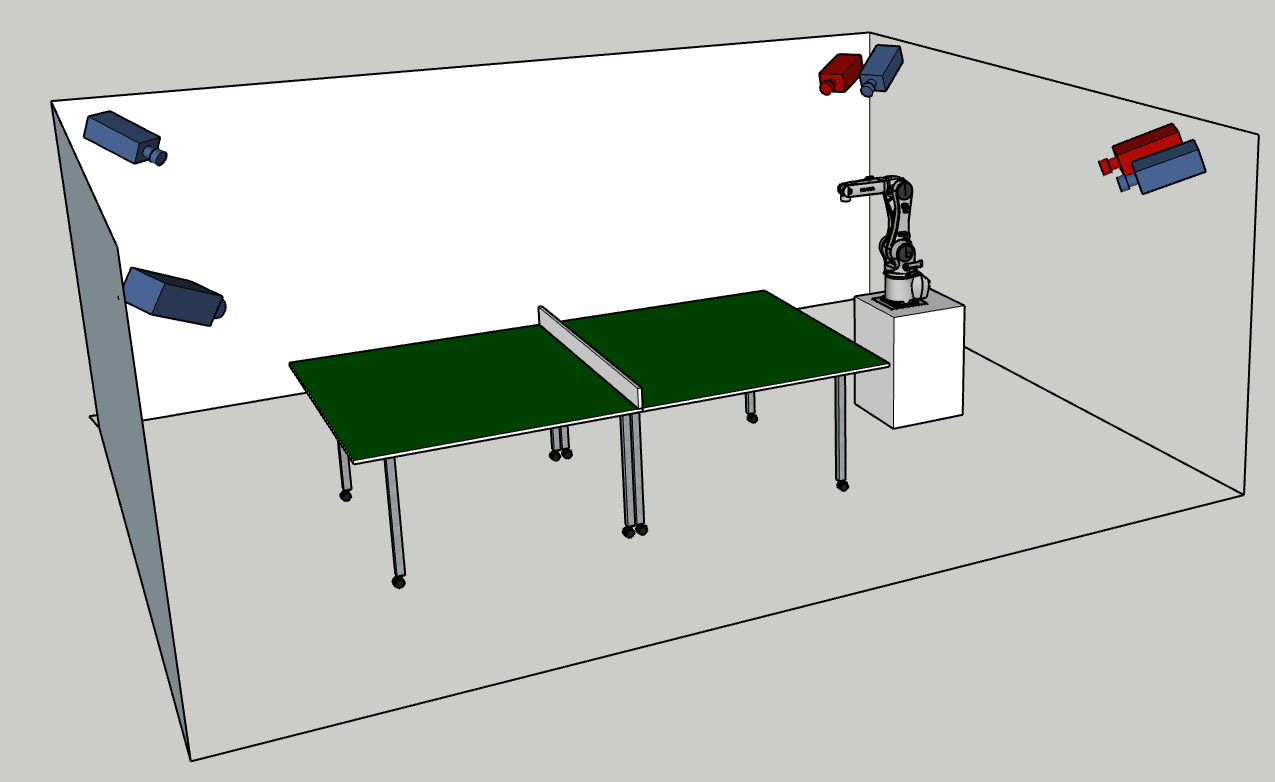}
    \caption{
        Our camera setup consists of {\color{blue}four frame-based cameras (in blue)} and {\color{red}two event-based cameras (in red)} with baselines of $3$m to $5$m.
        Only the event-based cameras are used in this work.
        Schematic is up to scale.
    }
    \label{fig:camera_setup}
\end{figure}

We used a table tennis robot setup similar to~\cite{Tebbe2019gcpr}, but extended it with two event-based cameras.
This camera system consists of two hardware-synchronized Prophesee EVK4 event-based cameras (1280x720 pixels).
Camera bias settings for the event-based cameras were configured to minimize noise and so that the flying ball would cause most events.
%
Additionally, the Event Trail Filter (STC and Trail)\footnote{https://docs.prophesee.ai/stable/hw/manuals/esp.html?highlight=stc\#event-trail-filter-stc-trail} were used to cancel redundant information.
The filter considers successive events of the same polarity generated in a short time as a burst and will remove the second event of a burst.
The whole table tennis robot system is visualized in \cref{fig:camera_setup}.
To calibrate this event-based camera system, we used the wand-based calibration approach introduced in~\cite{Gossard2023icra}.
A Butterfly Amicus Prime ball gun with default speed settings ($4$m/s) was used to shoot the table tennis balls from the opposite side of the robot arm.

For all the software components of our perception pipeline, we used ROS2~\cite{Macenski2022sr}.
We used the metavison\_driver\footnote{https://github.com/ros-event-camera/metavision\_driver} as ROS2 driver for the event-based cameras.

\subsection{The Perception Pipeline}\label{subsec:perception-pipeline}

%
%

\begin{figure}[t!]
    \centering
    \includegraphics[width=1.0\linewidth]{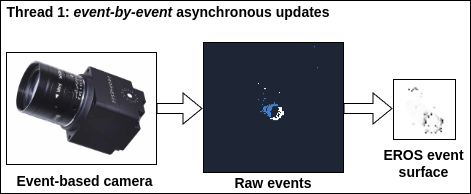}
    \caption{
        The first thread of our event-based perception pipeline.
        We process the incoming event and update the EROS event surface \textit{event-by-event}.
    }
    \label{fig:thread_1}
\end{figure}

Inspired by~\cite{Glover2022pami}, we use the best of processing events \textit{event-by-event} and in \textit{batches-of-events}.
We do this by running two threads simultaneously.
In the first thread, visualized in~\cref{fig:thread_1}, our perception pipeline processes the incoming events and updates the EROS event surface \textit{event-by-event}.
\begin{figure}[t!]
    \centering
    \includegraphics[width=0.97\linewidth]{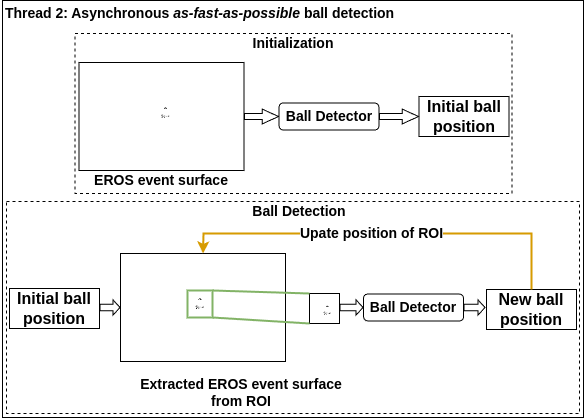}
    \caption{
        The second thread of our event-based perception pipeline.
        We initialize the \ac{ROI} of the ball and, afterward, detect the ball within the \ac{ROI} \textit{as-fast-as-possible}.
    }
    \label{fig:thread_2}
\end{figure}
In the second thread, shown in~\cref{fig:thread_2}, the ball detection runs \textit{as-fast-as-possible}.
In the next two sections, each of the two parts is described in more detail.

\subsection{Event-by-event update: EROS Surface}\label{subsec:event-update}

Our perception pipeline uses the asynchronously incoming events to update the EROS event representation \textit{event-by-event}.
The EROS event surface allows the decoupling of the asynchronous, high-temporal resolution event stream and the slower processing algorithm.

\begin{figure}[ht!]
    \centering
    \subfloat[]{
        \includegraphics[width=0.5\linewidth]{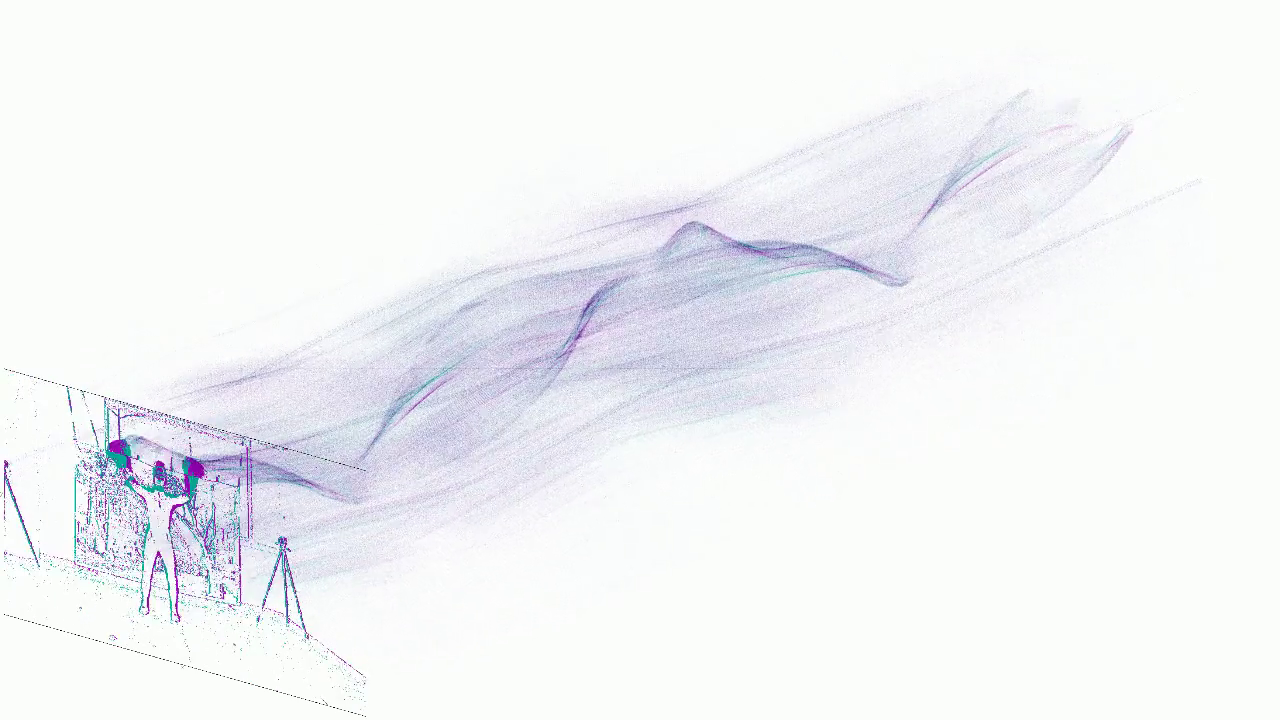}\label{fig:events_scene}
    }%
    \subfloat[]{
        \includegraphics[width=0.5\linewidth]{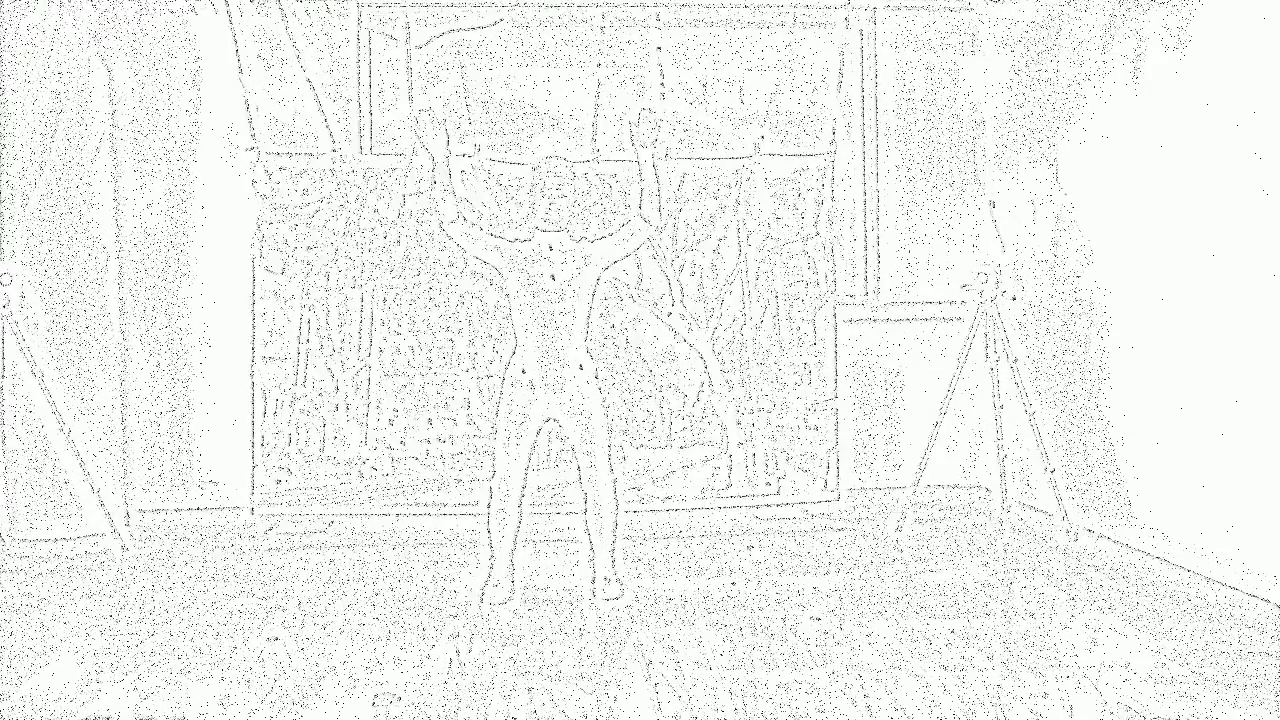}\label{fig:eros_scene}
    }%
    \caption{Example surface using the EROS algorithm to enable velocity-independent representation without temporal parameter tuning: (a) the input events over time for an example scene, (b) a visualization of the EROS surface after integrating event information over the entire time period of the dataset.}
    \label{fig:eros_surfaces}
\end{figure}

\begin{figure}[ht!]
    \centering
    \subfloat[]{
        \includegraphics[width=0.5\linewidth]{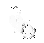}\label{fig:eros_ball}
    }%
    \subfloat[]{
        \includegraphics[width=0.5\linewidth]{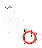}\label{fig:eros_ball_detected}
    }
    \caption{(a) A visualization of the EROS surface on the data captured for the perception pipeline of the table tennis robot.
            (b) The detected circle from the Hough-based circle detector overlayed.}
    \label{fig:ball_detection}
\end{figure}

EROS, with its update step defined in~\cref{alg:eros}, provides a coherent and bound spatial representation of the asynchronous events, partially maintaining the information about their temporal order and attempts to capture the most up-to-date position of edges in the scene~\cite{Glover2022pami}.
\Cref{fig:events_scene} shows the input events over time for an example scene.
The corresponding EROS event surface is shown in~\cref{fig:eros_scene}.
The EROS event surface of a flying table tennis ball, used in our application, is visualized in~\cref{fig:eros_ball}.

The process time of the EROS update depends on only one parameter, $k_{\text{EROS}}$, which corresponds to the update region size around each event at position $(v_x, v_y)$.
In this work, we use a $k_{\text{EROS}}$ of $10$.
Since the EROS update is a computationally cheap operation, the update can keep up with the incoming event stream up to approximately 10 Mevents/s~\cite{Glover2024icra}.
\begin{algorithm}[t!]
  \caption{Event-by-event EROS update}\label{alg:eros}
  \begin{algorithmic}
  \REQUIRE $d=0.3^{1.0 / k_{E R O S}}, k_{E R O S}, (v_x, v_y)$
  \FOR{$x=v_x-k_{E R O S}: v_x+k_{E R O S}$}
  \FOR{$y=v_y-k_{\text {EROS }}: v_y+k_{E R O S}$}
  \STATE $E R O S_{x y} \leftarrow E R O S_{x y} \cdot d$
  \ENDFOR
  \ENDFOR
  \STATE $\operatorname{EROS}_{v_x v_y} \leftarrow 255$
  \end{algorithmic}
\end{algorithm}


\subsection{As-fast-as-possible computation: Ball Detection}\label{subsec:ball-detection}

As previously mentioned, we use the EROS surface as an event representation.
The EROS surface can be sampled at any point in time with a temporal resolution of the event input and used as a ``grey image'', with values between $0$ and $255$.

Our ball detection is performed on the latest EROS surface, using a fast C++ implementation of a Hough-based circle detector, with an output similar to \textit{HoughCircles()} in OpenCV~\cite{Bradski2008djst}.
The velocity independence of EROS is critical for the Hough transform to succeed as blurred edges due to motion incorrectly bias Hough transform results.
An example of a detected ball is shown in~\cref{fig:eros_ball_detected}.

Ball detection is done in two modes: an initialization using the whole resolution and within a \acf{ROI} afterward.
Since the ball does not move far from one detection to the next, we use the previous ball position as the center of the following \ac{ROI}.
Restricting ball detections to this \ac{ROI} improves the runtime further, as shown in our experiments in~\cref{subsec:runtime}.

\begin{table*}[t]
  \caption{
    Overall results.
    $*$ Since the updates of the frame-based baseline are synchronous, in comparison to the asynchronous event-based pipelines, we report the result in Hz.
    $\bigtriangleup$The authors in~\cite{Tebbe2019gcpr} did not provide IoU results.
    $\dagger$The event-based median filter baseline only provides the position, and therefore, it is not possible to calculate the IoU.}
  \centering
  \begin{tabular}{
    @{}
    l
    S[table-number-alignment = left, exponent-mode = threshold, exponent-thresholds = 0:3, round-mode = uncertainty, round-precision = 2, round-pad = false]
    S[table-number-alignment = left]
    S[table-number-alignment = left]
    S[table-number-alignment = left]
    @{}
  }
    \hline
    {Method} & {Update rate [updates/s]} & {Average run time (initialization) [ms]} & {Error [pixels]} & IoU  \\
    \hline
    Frame-based baseline~\cite{Tebbe2019gcpr}
      & \num{149.0}\unit{\hertz}*  
      & 0.8 { (2.2)}
      & 1.32
      & {$\bigtriangleup$} \\
    Event-based median filter baseline 
      & \num{13750 \pm 1196}  
      & 0.00008 { (1.1)}
      & \num{2.90 \pm 1.60}
      & {$\dagger$} \\
    Event-based particle filter baseline~\cite{Glover2017iros} 
      & \num{3493 \pm 402}
      & 0.7 { (1.0)}
      & \num{2.93 \pm 1.13}
      & \num{0.55 \pm 0.08} \\
    Proposed event-based pipeline 
      & \num{4136.0 \pm 101}
      & 0.1 { (0.66)}
      & \num{1.34 \pm 0.79}
      & {\boldmath \num{0.78 \pm 0.11}} \\
    \hline
  \end{tabular}
  \label{tab:overall_results}
\end{table*}

\section{Experiments and Results}\label{sec:experiments}

Fast and accurate ball detection is crucial for enabling a table tennis robot to rally a ball back successfully.
The detected ball positions in both cameras need to be accurate so that the triangulated 3D ball positions are precise as well.
The speed and update rate of the ball detections are crucial since table tennis is a fast-paced game, and there is only between $0.1s$ and $1s$ time from the start of the ball's trajectory until the ball hits the robot's racket.
Accurate velocity and spin value estimates with a low uncertainty can additionally help with path planning and control of a robot arm.
Therefore, we designed our experiments to evaluate the mentioned key metrics for a perception pipeline used in a table tennis robot.

We start this experimental section by describing the baselines used in~\cref{subsec:baselines}.
Then, in our first experiment in~\cref{subsec:accuracy}, we measure the \textbf{accuracy} of the different perception pipelines.
Next, we measure and compare the \textbf{run time} of the perception pipelines in~\cref{subsec:runtime}.
Besides the run times, we are also interested in the \textbf{update rate} to know how frequently the pipelines report the position of the detected ball.
This is evaluated in~\cref{subsec:update_rate}.

We designed the proposed perception pipeline for a table tennis robot system, and therefore, it is crucial to highlight the benefits of the presented approach for the whole robot setup.
After the ball detection, the next step in a table tennis robot system is the prediction of the ball's trajectory.
In~\cref{subsec:trajectory_prediction}, we compare the trajectory prediction using the ball's 3D position from the frame-based and our proposed event-based system.

\subsection{The Baselines}\label{subsec:baselines}

In the following experiments, we compared our proposed perception pipeline with three other approaches that serve us as baselines.
The first is the frame-based system presented in~\cite{Tebbe2019gcpr}.
A first event-based perception baseline is a median filter approach, similar to~\cite{Delbruck2013fnins}.
The second event-based perception baseline is a particle filter based on~\cite{Glover2017iros}.

\paragraph{Frame-based pipeline}

Since the majority of the table tennis robot literature uses frame-based cameras for ball detection, we want to compare our proposed event-based method with one that uses only frame-based cameras.
While \ac{CNN} based ball detectors tend to have a slightly higher accuracy, they also tend to be slower than classical ball detectors, using more conventional image processing techniques.
%

We used the pipeline presented in~\cite{Tebbe2019gcpr}, which has state-of-the-art performance and used a similar setup.
In the first step, the method uses the difference between the current frame and one from the past to avoid being affected by static objects.
Afterward, metrics like the aspect ratio, circularity, and size are used to filter out outliers.
Since the frame-based cameras run at $149$fps, the update rate is $149$Hz with a processing latency of around $1$ms.

\paragraph{Event-based median filter}

Having static event-based cameras and assuming a static background, one of the simplest approaches to detecting a flying ball is to use the median of the events' positions in a time window.
A similar approach, called event clustering, was already used in~\cite{Delbruck2013fnins}.
This approach will fail if other moving objects in the scene trigger events.

To prevent this failure case to some degree, we used a \ac{ROI}, in which the median filter will consider events.
We implemented an initialization step to set the position of the first \ac{ROI}.
We do this initialization by using a blob detector on accumulated event frames, which we get from the \textit{event\_camera\_renderer}\footnote{https://github.com/ros-event-camera/event\_camera\_renderer} ROS2 node.
Later on, the \ac{ROI} is set to the position of the latest detected ball.

\paragraph{Event-based particle filter}

A more sophisticated, state-of-the-art approach, presented in~\cite{Glover2017iros}, uses a particle filter.
The particle filter not only estimates the ball's position and radius but also the time window for the next step.
This allows the approach to adapt to the speed of the moving ball and, therefore, to the difference in the resulting event rate.
While we started from the original implementation of~\cite{Glover2017iros}, we adjusted the implementation and the parameters to work with the fast movement of our flying table tennis balls.

For this particle filter approach, we use the same initialization step as for the median filter described above.

\subsection{Accuracy}\label{subsec:accuracy}

To measure the accuracy of the event-based perception pipelines, we proceeded as follows.
We compared the 2D position estimates of all the approaches from multiple ball trajectories with ground truth positions.
To get ground truth 2D positions, we used a circle detector on reconstructed frames and manually removed inaccurate detections.
We used e2vid~\cite{Rebecq2019pami} for the reconstructed frames with our event stream.
This resulted in $286$ ground truth 2D positions.

The 2D ball detections of the different event-based vision pipelines will not have the same time stamps due to the event data's asynchronous nature and the different approaches' working principles.
To account for this, we linearly interpolated the 2D positions to match with the time stamps of the ground truth positions.
Given the interpolated 2D positions and the ground truth positions, we calculated the pixel error between them.
The results are listed in the column ``Error'' in~\cref{tab:overall_results}.

As can be seen, the frame-based pipeline does achieve the lowest pixel error with $1.32$ pixels, although the proposed event-based perception pipeline with an error of $1.34$ pixels is very close.
The event-based baselines have an error of $2.90$ and $2.93$ pixels, respectively, an error more than double the one of the proposed pipeline.
The proposed pipeline also has a lower standard deviation ($\pm0.79$ pixels) than the event-based baselines ($\pm1.60$ and $\pm1.13$ pixels), indicating more stable 2D ball detections.

\subsection{Runtime}\label{subsec:runtime}

To measure the run times of the ball detection pipelines and their components, we used the time measurement functionality of the C++ standard library.
We report the mean and standard deviation of two ball trajectories.
Since the different ball detection pipelines have various software components, we list the run times for every pipeline separately.

The results of the frame-based baseline are listed in~\cref{tab:runtime_frame}.
In~\cref{tab:runtime_events_median}, we list the run times of the event-based median filter approach.
The results of the event-based particle filter approach are listed in~\cref{tab:runtime_events_pf}.
For our proposed event-based pipeline, we list the run times in~\cref{tab:runtime_events_eros}.
The overall average run times are listed in~\cref{tab:overall_results}.
\begin{table}[ht!]
  \caption{Run times of the frame-based baseline~\cite{Tebbe2019gcpr}}
  \centering
  \begin{tabular}{
    @{}
    l
    S[table-number-alignment = right, exponent-mode = threshold, exponent-thresholds = -3:3, round-mode = uncertainty, round-precision = 2, round-pad = false]
    @{}
  }
  \hline
  {Module} & {Runtime [$\mu$s]} \\
  \hline
  \hline
  Ball detection (full resolution) & 2200.0 \\
  Ball detection (\ac{ROI}) & 800.0 \\
  \hline
  \end{tabular}
  \label{tab:runtime_frame}
\end{table}
\begin{table}[ht!]
  \caption{Run times of the event-based median filter baseline}
  \centering
  \begin{tabular}{
    @{}
    l
    S[table-number-alignment = right, exponent-mode = threshold, exponent-thresholds = -3:3, round-mode = uncertainty, round-precision = 2, round-pad = false]
    @{}
  }
  \hline
  {Module} & {Runtime [$\mu$s]} \\
  \hline
  \hline
  Initialization & \num{1059 \pm 186.68} \\
  Update & \num{0.08 \pm 0.62} \\
  \hline
  \end{tabular}
  \label{tab:runtime_events_median}
\end{table}
\begin{table}[ht!]
  \caption{Run times of the event-based particle filter baseline~\cite{Glover2017iros}}
  \centering
  \begin{tabular}{
    @{}
    l
    S[table-number-alignment = right, exponent-mode = threshold, exponent-thresholds = -3:3, round-mode = uncertainty, round-precision = 2, round-pad = false]
    @{}
  }
  \hline
  {Module} & {Runtime [$\mu$s]} \\
  \hline
  \hline
  Initialization & \num{1045.07 \pm 203.88} \\
  Fetch events & \num{5.26 \pm 10.52} \\
  Particle filter update & \num{698.56 \pm 727.15} \\
  Maximum likelihood calculation & \num{0.19 \pm 0.42} \\
  \hline
  \end{tabular}
  \label{tab:runtime_events_pf}
\end{table}
\begin{table}[ht!]
  \caption{Run times of our proposed event-based pipeline}
  \centering
  \begin{tabular}{
    @{}
    l
    S[table-number-alignment = right, exponent-mode = threshold, exponent-thresholds = -3:3, round-mode = uncertainty, round-precision = 2, round-pad = false]
    @{}
  }
  \hline
  {Module} & {Runtime [$\mu$s]} \\
  \hline
  \hline
  Fetch EROS surface & \num{0.47 \pm 1.27} \\
  Ball detection (full resolution) & \num{655.94 \pm 37.62} \\
  Ball detection (\ac{ROI}) & \num{90.74 \pm 17.76} \\
  \hline
  \end{tabular}
  \label{tab:runtime_events_eros}
\end{table}

As can be seen in~\cref{tab:overall_results}, the event-based particle filter approach has a similar average run time as the frame-based approach.
In contrast, the other two event-based approaches are significantly faster.
During the initialization, the event-based approaches have an average run time of at most half of the frame-based baseline.

As previously mentioned, the median- and particle-filter share the same initialization schema, and therefore, both have an average initialization run time of around $1.1$ms.
With an average update run time of $0.08\mu$s, the event-based median filter has the lowest run time after the initialization.

For the event-based particle filter method, the update step is, with around $700\mu$s, the most costly component.
Fetching the events ($5\mu$s) and the maximum likelihood calculation ($0.19\mu$s) is negligible compared to the update step.

In our proposed event-based perception pipeline, the ball detection takes an average of $656\mu$s during the initialization and $91\mu$s in the \ac{ROI}.
This is a bigger reduction compared to the frame-based baseline.
Fetching the EROS time surface is with $0.5\mu$s negligible.

\subsection{Update Rate}\label{subsec:update_rate}

While a fast run time is desirable since it reduces the latency of the algorithms, the update rate is also of great interest.
In other words, we not only want to have the position estimates as soon as possible, but we also want them as frequently as possible.

To measure the update rates, we divided the number of position updates by the duration of the trajectory, $$\text{update rate} = \frac{\# \text{position updates}}{\text{duration}}.$$
We measured the durations of the ball trajectories manually by visualizing the event stream in Metavision Studio\footnote{https://docs.prophesee.ai/stable/metavision\_studio}.
We did this for four different trajectories and report the mean and standard deviation in~\cref{tab:overall_results} in the column ``Update rate''.

As listed in~\cref{tab:overall_results}, the frame-based baseline has with $149$Hz the lowest update rate, which is dictated by the frame rate of the frame-based cameras.
The event-based median filter has the highest update rate, with an average of $1.38\cdot10^4\frac{\text{updates}}{s}$, which comes from the fact that it also has the lowest run time.
The event-based particle filter ($3.49\cdot10^3\frac{\text{updates}}{s}$) and the proposed event-based perception pipeline ($4.14\cdot10^3\frac{\text{updates}}{s}$) are quite close together, with the latter being superior.

This experiment shows the benefits of an event-based perception pipeline.
The update rate is not restricted by the sensor but by the algorithm used, which processes the event data.

\subsection{Trajectory Prediction}\label{subsec:trajectory_prediction}

After having detected the table tennis ball with both event-based cameras, the ball's 3D position can be triangulated, as sketched in~\cref{fig:eye-catcher}.
These 3D positions can then be used to estimate the ball's trajectory.
The more accurate the ball's predicted trajectory is, the better the robot arm can hit back.
In this experiment, we used the \acf{EKF}-based approach presented in~\cite{Tebbe2019gcpr} for the trajectory prediction.
Next to the position, this \ac{EKF} implementation also estimates the ball's velocity and spin.

More frequent 3D ball position updates can help the \ac{EKF} to reduce its uncertainties of the position, velocity, and spin of the flying ball.
As previous work discussed, the spin of the ball is crucial for a table tennis robot~\cite{Tebbe2019gcpr}\cite{Gossard2023iros}\cite{Gossard2024cvprw}.
Therefore, having a lower uncertainty about the ball's spin is beneficial.

To show the benefits of our proposed event-based perception pipeline, we predicted the ball's trajectory with the mentioned \ac{EKF}-based approach.
We do this by providing the \ac{EKF} the ball's 3D positions in the measurement update step and letting the \ac{EKF} predict the next position, velocity, and spin.
For a fair comparison, we estimated the transition covariance, the observation covariance, the initial state means, and the covariance of the initial state for both \acp{EKF} using the \ac{EM} algorithm and made the predictions with these optimized \ac{EKF} parameters.
The \ac{EM} algorithm optimizes the log-likelihood of the predicted states and, therefore, adjusts the parameters of the \ac{EKF} to the data.
This ensures good parameters for both \ac{EKF}-based trajectory predictors (frame-based and event-based).
Our implementation of \ac{EKF} and the \ac{EM}-based parameter optimization is based on pykalman\footnote{https://github.com/pykalman/pykalman} which we adjusted to work with the non-linear model of the ball’s trajectory mid-air, commonly used in the literature~\cite{Zhang2010tim}~\cite{Tebbe2019gcpr}.

We measure the error between the predicted ball position and the ball position from the triangulated trajectory.
Since we do not have observations of the spin and velocity, we emphasize the uncertainties of the estimated velocities and spin values given by the \ac{EKF}.

\begin{figure}[t!]
    \centering
    \includegraphics[width=1.0\linewidth]{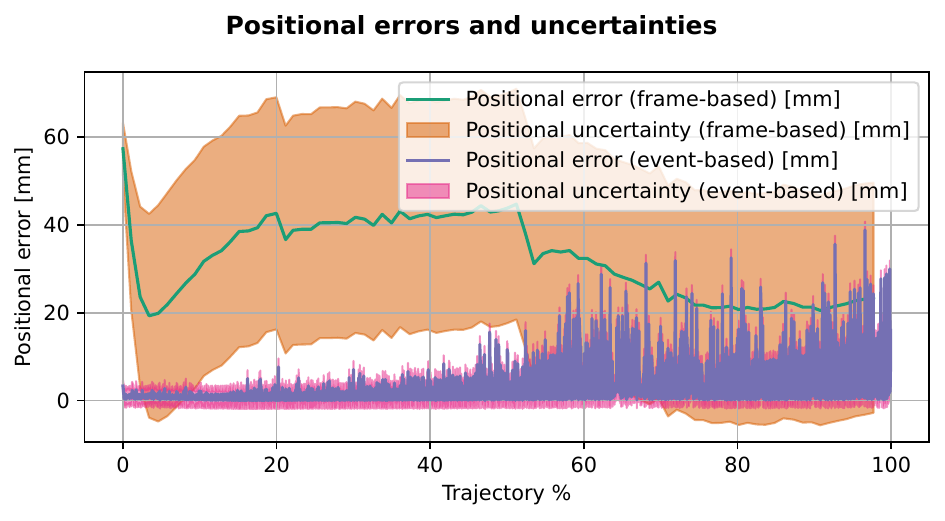}
    \caption{
      The absolute positional error and the positional uncertainty of the trajectory with the proposed event-based perception pipeline and with the frame-based baseline.
    }
    \label{fig:positional-uncertainties}
\end{figure}
\begin{figure}[t!]
    \centering
    \includegraphics[width=1.0\linewidth]{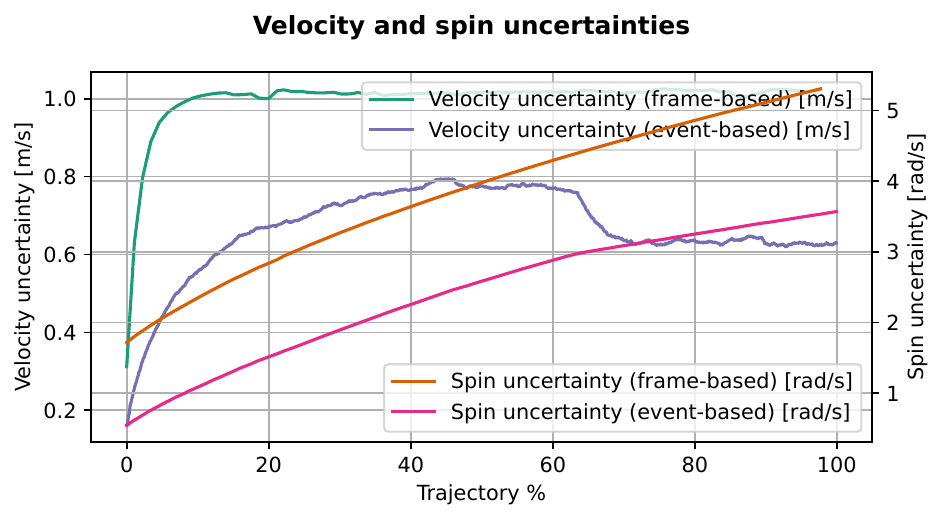}
    \caption{
      The uncertainty of the predicted velocity and the uncertainty of the predicted spin of the trajectory with the proposed event-based perception pipeline and with the frame-based baseline.
    }
    \label{fig:velocity-spin-uncertainties}
\end{figure}

In~\cref{fig:positional-uncertainties}, we visualized the positional errors and the corresponding uncertainties of our event-based pipeline and of the frame-based baseline.
In~\cref{fig:velocity-spin-uncertainties}, we show the uncertainties of the velocity and spin estimates of our event-based pipeline and of the frame-based baseline.

As we can see in~\cref{fig:positional-uncertainties}, the positional error of the event-based pipeline drops faster at the start of the trajectory.
We can also see that the positional error of the event-based pipeline is lower compared to the frame-based baseline most of the time.
The difference in the course of the positional error between the frame-based and the event-based pipeline is partially influenced by the different camera setups.
As sketched in~\cref{fig:camera_setup}, the two event-based cameras are located on the side of the robot, whereas the frame-based cameras are placed on all four corners of the ceiling.

Similar to the positional error, the uncertainties of the event-based pipeline are also smaller compared to the frame-based baseline, most of the time.

In~\cref{fig:velocity-spin-uncertainties}, we can see that the uncertainties of both the velocity and the spin estimates are lower when using the event-based pipeline.
%
This showcases the benefit of using an asynchronous event-based perception pipeline instead of a frame-based one.

\section{Conclusion}

Table tennis robots are a challenging research field for robotic perception algorithms and, therefore, have gained traction over the last few years.
To the best of our knowledge, we present the first real-time event-based perception pipeline for a table tennis robot.
While not yet integrated into a whole table tennis robot system, using a table tennis robot setup, we showcase the benefits of an event-based perception pipeline over perception pipelines that use traditional frame-based cameras.
Namely, an update rate that is an order of magnitude faster than a frame-based perception pipeline while having a comparable accuracy.
Furthermore, we show that the higher update rate results in lower uncertainties of the estimated position, velocity, and spin.
This is beneficial for the path planning and robot control of a table tennis robot, which has to be fast to rally the ball back successfully.

We hope this work will motivate more researchers working on table tennis robots to use event-based computer vision.

\section*{Acknowledgments}
Special thanks to Sony AI for partially funding this project.
We would also like to thank Mario Laux for his help with the C++ implementation of the ball detector.
Thanks to Bernd Pfrommer for his open source contributions to many (event-based vision) ROS packages, two of them used in this work.

\bibliographystyle{IEEEtran}
\bibliography{IEEEabrv,bibliography.bib}

\begin{thebibliography}{10}
\providecommand{\url}[1]{#1}
\csname url@samestyle\endcsname
\providecommand{\newblock}{\relax}
\providecommand{\bibinfo}[2]{#2}
\providecommand{\BIBentrySTDinterwordspacing}{\spaceskip=0pt\relax}
\providecommand{\BIBentryALTinterwordstretchfactor}{4}
\providecommand{\BIBentryALTinterwordspacing}{\spaceskip=\fontdimen2\font plus
\BIBentryALTinterwordstretchfactor\fontdimen3\font minus
  \fontdimen4\font\relax}
\providecommand{\BIBforeignlanguage}[2]{{%
\expandafter\ifx\csname l@#1\endcsname\relax
\typeout{** WARNING: IEEEtran.bst: No hyphenation pattern has been}%
\typeout{** loaded for the language `#1'. Using the pattern for}%
\typeout{** the default language instead.}%
\else
\language=\csname l@#1\endcsname
\fi
#2}}
\providecommand{\BIBdecl}{\relax}
\BIBdecl

\bibitem{Ziegler2023corlw}
\BIBentryALTinterwordspacing
A.~Ziegler, T.~Gossard, K.~Vetter, J.~Tebbe, and A.~Zell, ``A multi-modal table
  tennis robot system,'' in \emph{RoboLetics: Workshop on Robot Learning in
  Athletics @CoRL 2023}, 2023. [Online]. Available:
  \url{https://arxiv.org/abs/2310.19062}
\BIBentrySTDinterwordspacing

\bibitem{Ambrosio2024arxiv}
\BIBentryALTinterwordspacing
D.~B. D'Ambrosio, S.~Abeyruwan, L.~Graesser, A.~Iscen, H.~B. Amor, A.~Bewley,
  B.~J. Reed, K.~Reymann, L.~Takayama, Y.~Tassa, K.~Choromanski, E.~Coumans,
  D.~Jain, N.~Jaitly, N.~Jaques, S.~Kataoka, Y.~Kuang, N.~Lazic, R.~Mahjourian,
  S.~Moore, K.~Oslund, A.~Shankar, V.~Sindhwani, V.~Vanhoucke, G.~Vesom, P.~Xu,
  and P.~R. Sanketi, ``Achieving human level competitive robot table tennis,''
  2024. [Online]. Available: \url{https://arxiv.org/abs/2408.03906}
\BIBentrySTDinterwordspacing

\bibitem{Tebbe2019gcpr}
\BIBentryALTinterwordspacing
J.~Tebbe, Y.~Gao, M.~Sastre-Rienietz, and A.~Zell, ``A table tennis robot
  system using an industrial {KUKA} robot arm,'' in \emph{Lecture Notes in
  Computer Science}.\hskip 1em plus 0.5em minus 0.4em\relax Springer
  International Publishing, 2019, pp. 33--45. [Online]. Available:
  \url{https://doi.org/10.1007/978-3-030-12939-2_3}
\BIBentrySTDinterwordspacing

\bibitem{DAmbrosio2023rss}
\BIBentryALTinterwordspacing
D.~D{\textquotesingle}Ambrosio, N.~Jaitly, V.~Sindhwani, K.~Oslund, P.~Xu,
  N.~Lazic, A.~Shankar, T.~Ding, J.~Abelian, E.~Coumans, G.~Kouretas,
  T.~Nguyen, J.~Boyd, A.~Iscen, R.~Mahjourian, V.~Vanhoucke, A.~Bewley,
  Y.~Kuang, M.~Ahn, D.~Jain, S.~Kataoka, O.~Cortes, P.~Sermanet, C.~Lynch,
  P.~Sanketi, K.~Choromanski, W.~Gao, J.~Kangaspunta, K.~Reymann, G.~Vesom,
  S.~Moore, A.~Singh, S.~Abeyruwan, and L.~Graesser, ``Robotic table tennis: A
  case study into a high speed learning system,'' in \emph{Robotics: Science
  and Systems {XIX}}.\hskip 1em plus 0.5em minus 0.4em\relax Robotics: Science
  and Systems Foundation, Jul. 2023. [Online]. Available:
  \url{https://doi.org/10.15607/rss.2023.xix.006}
\BIBentrySTDinterwordspacing

\bibitem{GomezGonzalez2019robotics}
\BIBentryALTinterwordspacing
S.~Gomez-Gonzalez, Y.~Nemmour, B.~Sch\"{o}lkopf, and J.~Peters, ``Reliable
  real-time ball tracking for robot table tennis,'' \emph{Robotics}, vol.~8,
  no.~4, p.~90, Oct. 2019. [Online]. Available:
  \url{https://doi.org/10.3390/robotics8040090}
\BIBentrySTDinterwordspacing

\bibitem{Ding2022iros}
\BIBentryALTinterwordspacing
T.~Ding, L.~Graesser, S.~Abeyruwan, D.~B. D{\textquotesingle}Ambrosio,
  A.~Shankar, P.~Sermanet, P.~R. Sanketi, and C.~Lynch, ``Learning high speed
  precision table tennis on a physical robot,'' in \emph{2022 {IEEE}/{RSJ}
  International Conference on Intelligent Robots and Systems ({IROS})}.\hskip
  1em plus 0.5em minus 0.4em\relax {IEEE}, Oct. 2022. [Online]. Available:
  \url{https://doi.org/10.1109/iros47612.2022.9982205}
\BIBentrySTDinterwordspacing

\bibitem{Gallego2020pami}
G.~Gallego, T.~Delbrück, G.~Orchard, C.~Bartolozzi, B.~Taba, A.~Censi,
  S.~Leutenegger, A.~J. Davison, J.~Conradt, K.~Daniilidis, and D.~Scaramuzza,
  ``Event-based vision: A survey,'' \emph{IEEE Transactions on Pattern Analysis
  and Machine Intelligence}, vol.~44, no.~1, pp. 154--180, 2022.

\bibitem{Monforte2020aicas}
\BIBentryALTinterwordspacing
M.~Monforte, A.~Arriandiaga, A.~Glover, and C.~Bartolozzi, ``Exploiting event
  cameras for spatio-temporal prediction of fast-changing trajectories,'' in
  \emph{2020 2nd IEEE International Conference on Artificial Intelligence
  Circuits and Systems (AICAS)}.\hskip 1em plus 0.5em minus 0.4em\relax IEEE,
  Aug. 2020. [Online]. Available:
  \url{http://dx.doi.org/10.1109/AICAS48895.2020.9073855}
\BIBentrySTDinterwordspacing

\bibitem{Mitrokhin2018iros}
\BIBentryALTinterwordspacing
A.~Mitrokhin, C.~Fermuller, C.~Parameshwara, and Y.~Aloimonos, ``Event-based
  moving object detection and tracking,'' in \emph{2018 IEEE/RSJ International
  Conference on Intelligent Robots and Systems (IROS)}.\hskip 1em plus 0.5em
  minus 0.4em\relax IEEE, Oct. 2018. [Online]. Available:
  \url{http://dx.doi.org/10.1109/IROS.2018.8593805}
\BIBentrySTDinterwordspacing

\bibitem{Forrai2023icra}
\BIBentryALTinterwordspacing
B.~Forrai, T.~Miki, D.~Gehrig, M.~Hutter, and D.~Scaramuzza, ``Event-based
  agile object catching with a quadrupedal robot,'' in \emph{2023 IEEE
  International Conference on Robotics and Automation (ICRA)}.\hskip 1em plus
  0.5em minus 0.4em\relax IEEE, May 2023. [Online]. Available:
  \url{http://dx.doi.org/10.1109/ICRA48891.2023.10161392}
\BIBentrySTDinterwordspacing

\bibitem{Perot2020neurips}
\BIBentryALTinterwordspacing
E.~Perot, P.~de~Tournemire, D.~Nitti, J.~Masci, and A.~Sironi, ``Learning to
  detect objects with a 1 megapixel event camera,'' in \emph{Advances in Neural
  Information Processing Systems}, H.~Larochelle, M.~Ranzato, R.~Hadsell, M.~F.
  Balcan, and H.~Lin, Eds., vol.~33.\hskip 1em plus 0.5em minus 0.4em\relax
  Curran Associates, Inc., 2020, pp. 16\,639--16\,652. [Online]. Available:
  \url{https://proceedings.neurips.cc/paper/2020/file/c213877427b46fa96cff6c39e837ccee-Paper.pdf}
\BIBentrySTDinterwordspacing

\bibitem{Stoffregen2020eccv}
\BIBentryALTinterwordspacing
T.~Stoffregen, C.~Scheerlinck, D.~Scaramuzza, T.~Drummond, N.~Barnes,
  L.~Kleeman, and R.~Mahony, \emph{Reducing the Sim-to-Real Gap for Event
  Cameras}.\hskip 1em plus 0.5em minus 0.4em\relax Springer International
  Publishing, 2020, p. 534–549. [Online]. Available:
  \url{http://dx.doi.org/10.1007/978-3-030-58583-9_32}
\BIBentrySTDinterwordspacing

\bibitem{Glover2022pami}
\BIBentryALTinterwordspacing
A.~Glover, A.~Dinale, L.~D.~S. Rosa, S.~Bamford, and C.~Bartolozzi,
  ``luvharris: A practical corner detector for event-cameras,'' \emph{IEEE
  Transactions on Pattern Analysis and Machine Intelligence}, vol.~44, no.~12,
  p. 10087–10098, Dec. 2022. [Online]. Available:
  \url{http://dx.doi.org/10.1109/TPAMI.2021.3135635}
\BIBentrySTDinterwordspacing

\bibitem{Glover2024icra}
\BIBentryALTinterwordspacing
A.~Glover, L.~Gava, Z.~Li, and C.~Bartolozzi, ``Edopt: Event-camera 6-dof
  dynamic object pose tracking,'' in \emph{2024 IEEE International Conference
  on Robotics and Automation (ICRA)}.\hskip 1em plus 0.5em minus 0.4em\relax
  IEEE, May 2024, p. 18200–18206. [Online]. Available:
  \url{http://dx.doi.org/10.1109/ICRA57147.2024.10611511}
\BIBentrySTDinterwordspacing

\bibitem{Billingsley1983robot}
J.~Billingsley, ``Robot ping pong,'' \emph{Practical Computing}, vol.~6, no.~5,
  1983.

\bibitem{Buchler2022tor}
\BIBentryALTinterwordspacing
D.~Buchler, S.~Guist, R.~Calandra, V.~Berenz, B.~Scholkopf, and J.~Peters,
  ``Learning to play table tennis from scratch using muscular robots,''
  \emph{{IEEE} Transactions on Robotics}, vol.~38, no.~6, pp. 3850--3860, Dec.
  2022. [Online]. Available: \url{https://doi.org/10.1109/tro.2022.3176207}
\BIBentrySTDinterwordspacing

\bibitem{Ji2021iros}
\BIBentryALTinterwordspacing
Y.~Ji, X.~Hu, Y.~Chen, Y.~Mao, G.~Wang, Q.~Li, and J.~Zhang, ``Model-based
  trajectory prediction and hitting velocity control for a new table tennis
  robot,'' in \emph{2021 IEEE/RSJ International Conference on Intelligent
  Robots and Systems (IROS)}.\hskip 1em plus 0.5em minus 0.4em\relax IEEE, Sep.
  2021. [Online]. Available:
  \url{http://dx.doi.org/10.1109/IROS51168.2021.9636000}
\BIBentrySTDinterwordspacing

\bibitem{Yang2021ieee}
\BIBentryALTinterwordspacing
L.~Yang, H.~Zhang, X.~Zhu, and X.~Sheng, ``Ball motion control in the table
  tennis robot system using time-series deep reinforcement learning,''
  \emph{IEEE Access}, vol.~9, p. 99816–99827, 2021. [Online]. Available:
  \url{http://dx.doi.org/10.1109/ACCESS.2021.3093340}
\BIBentrySTDinterwordspacing

\bibitem{Gao2020iros}
\BIBentryALTinterwordspacing
W.~Gao, L.~Graesser, K.~Choromanski, X.~Song, N.~Lazic, P.~Sanketi,
  V.~Sindhwani, and N.~Jaitly, ``Robotic table tennis with model-free
  reinforcement learning,'' in \emph{2020 {IEEE}/{RSJ} International Conference
  on Intelligent Robots and Systems ({IROS})}.\hskip 1em plus 0.5em minus
  0.4em\relax {IEEE}, Oct. 2020. [Online]. Available:
  \url{https://doi.org/10.1109/iros45743.2020.9341191}
\BIBentrySTDinterwordspacing

\bibitem{Gao2022ijcnn}
\BIBentryALTinterwordspacing
Y.~Gao, J.~Tebbe, and A.~Zell, ``A model-free approach to stroke learning for
  robotic table tennis,'' in \emph{2022 International Joint Conference on
  Neural Networks ({IJCNN})}.\hskip 1em plus 0.5em minus 0.4em\relax {IEEE},
  Jul. 2022. [Online]. Available:
  \url{https://doi.org/10.1109/ijcnn55064.2022.9892776}
\BIBentrySTDinterwordspacing

\bibitem{Achterhold2023l4dc}
J.~Achterhold, P.~Tobuschat, H.~Ma, D.~B{\"u}chler, M.~Muehlebach, and
  J.~Stueckler, ``Black-box vs. gray-box: A case study on learning table tennis
  ball trajectory prediction with spin and impacts,'' in \emph{Proceedings of
  the Learning for Dynamics and Control Conference (L4DC)}, 2023.

\bibitem{Cannici2019cvprw}
\BIBentryALTinterwordspacing
M.~Cannici, M.~Ciccone, A.~Romanoni, and M.~Matteucci, ``Asynchronous
  convolutional networks for object detection in neuromorphic cameras,'' in
  \emph{2019 {IEEE}/{CVF} Conference on Computer Vision and Pattern Recognition
  Workshops ({CVPRW})}.\hskip 1em plus 0.5em minus 0.4em\relax {IEEE}, Jun.
  2019. [Online]. Available: \url{https://doi.org/10.1109/cvprw.2019.00209}
\BIBentrySTDinterwordspacing

\bibitem{Schaefer2022cvpr}
\BIBentryALTinterwordspacing
S.~Schaefer, D.~Gehrig, and D.~Scaramuzza, ``{AEGNN}: Asynchronous event-based
  graph neural networks,'' in \emph{2022 {IEEE}/{CVF} Conference on Computer
  Vision and Pattern Recognition ({CVPR})}.\hskip 1em plus 0.5em minus
  0.4em\relax {IEEE}, Jun. 2022. [Online]. Available:
  \url{https://doi.org/10.1109/cvpr52688.2022.01205}
\BIBentrySTDinterwordspacing

\bibitem{Liu2024wacv}
Y.~Liu, M.~Gehrig, N.~Messikommer, M.~Cannici, and D.~Scaramuzza, ``Revisiting
  token pruning for object detection and instance segmentation,'' in
  \emph{Proceedings of the IEEE/CVF Winter Conference on Applications of
  Computer Vision (WACV)}, 2024.

\bibitem{Glover2016iros}
\BIBentryALTinterwordspacing
A.~Glover and C.~Bartolozzi, ``Event-driven ball detection and gaze fixation in
  clutter,'' in \emph{2016 {IEEE}/{RSJ} International Conference on Intelligent
  Robots and Systems ({IROS})}.\hskip 1em plus 0.5em minus 0.4em\relax {IEEE},
  Oct. 2016. [Online]. Available:
  \url{https://doi.org/10.1109/iros.2016.7759345}
\BIBentrySTDinterwordspacing

\bibitem{Glover2017iros}
\BIBentryALTinterwordspacing
------, ``Robust visual tracking with a freely-moving event camera,'' in
  \emph{2017 IEEE/RSJ International Conference on Intelligent Robots and
  Systems (IROS)}.\hskip 1em plus 0.5em minus 0.4em\relax {IEEE}, Sep. 2017.
  [Online]. Available: \url{http://dx.doi.org/10.1109/iros.2017.8206226}
\BIBentrySTDinterwordspacing

\bibitem{Falanga2020sr}
\BIBentryALTinterwordspacing
D.~Falanga, K.~Kleber, and D.~Scaramuzza, ``Dynamic obstacle avoidance for
  quadrotors with event cameras,'' \emph{Science Robotics}, vol.~5, no.~40,
  Mar. 2020. [Online]. Available:
  \url{http://dx.doi.org/10.1126/scirobotics.aaz9712}
\BIBentrySTDinterwordspacing

\bibitem{Scheerlinck2019accv}
\BIBentryALTinterwordspacing
C.~Scheerlinck, N.~Barnes, and R.~Mahony, \emph{Continuous-Time Intensity
  Estimation Using Event Cameras}.\hskip 1em plus 0.5em minus 0.4em\relax
  Springer International Publishing, 2019, p. 308–324. [Online]. Available:
  \url{http://dx.doi.org/10.1007/978-3-030-20873-8_20}
\BIBentrySTDinterwordspacing

\bibitem{Seifozzakerini2016bmcv}
\BIBentryALTinterwordspacing
S.~Seifozzakerini, W.-Y. Yau, B.~Zhao, and K.~Mao, ``Event-based hough
  transform in a spiking neural network for multiple line detection and
  tracking using a dynamic vision sensor,'' in \emph{Procedings of the British
  Machine Vision Conference 2016}, ser. BMVC 2016.\hskip 1em plus 0.5em minus
  0.4em\relax British Machine Vision Association, 2016, pp. 94.1--94.12.
  [Online]. Available: \url{http://dx.doi.org/10.5244/C.30.94}
\BIBentrySTDinterwordspacing

\bibitem{Lagorce2017pami}
\BIBentryALTinterwordspacing
X.~Lagorce, G.~Orchard, F.~Galluppi, B.~E. Shi, and R.~B. Benosman, ``Hots: A
  hierarchy of event-based time-surfaces for pattern recognition,'' \emph{IEEE
  Transactions on Pattern Analysis and Machine Intelligence}, vol.~39, no.~7,
  p. 1346–1359, Jul. 2017. [Online]. Available:
  \url{http://dx.doi.org/10.1109/TPAMI.2016.2574707}
\BIBentrySTDinterwordspacing

\bibitem{Sironi2018cvpr}
\BIBentryALTinterwordspacing
A.~Sironi, M.~Brambilla, N.~Bourdis, X.~Lagorce, and R.~Benosman, ``Hats:
  Histograms of averaged time surfaces for robust event-based object
  classification,'' in \emph{2018 IEEE/CVF Conference on Computer Vision and
  Pattern Recognition}.\hskip 1em plus 0.5em minus 0.4em\relax IEEE, Jun. 2018.
  [Online]. Available: \url{http://dx.doi.org/10.1109/CVPR.2018.00186}
\BIBentrySTDinterwordspacing

\bibitem{Gehrig2019iccv}
\BIBentryALTinterwordspacing
D.~Gehrig, A.~Loquercio, K.~Derpanis, and D.~Scaramuzza, ``End-to-end learning
  of representations for asynchronous event-based data,'' in \emph{2019
  IEEE/CVF International Conference on Computer Vision (ICCV)}.\hskip 1em plus
  0.5em minus 0.4em\relax IEEE, Oct. 2019, p. 5632–5642. [Online]. Available:
  \url{http://dx.doi.org/10.1109/ICCV.2019.00573}
\BIBentrySTDinterwordspacing

\bibitem{Gossard2023icra}
\BIBentryALTinterwordspacing
T.~Gossard, A.~Ziegler, L.~Kolmar, J.~Tebbe, and A.~Zell, ``ewand: A
  calibration framework for wide baseline frame-based and event-based camera
  systems,'' in \emph{2024 {International} {Conference} on {Robotics} and
  {Automation} ({ICRA})}.\hskip 1em plus 0.5em minus 0.4em\relax IEEE, 2024.
  [Online]. Available: \url{https://arxiv.org/pdf/2309.12685.pdf}
\BIBentrySTDinterwordspacing

\bibitem{Macenski2022sr}
\BIBentryALTinterwordspacing
S.~Macenski, T.~Foote, B.~Gerkey, C.~Lalancette, and W.~Woodall, ``Robot
  operating system 2: Design, architecture, and uses in the wild,''
  \emph{Science Robotics}, vol.~7, no.~66, p. eabm6074, 2022. [Online].
  Available: \url{https://www.science.org/doi/abs/10.1126/scirobotics.abm6074}
\BIBentrySTDinterwordspacing

\bibitem{Bradski2008djst}
G.~Bradski, ``{The OpenCV Library},'' \emph{Dr. Dobb's Journal of Software
  Tools}, 2000.

\bibitem{Delbruck2013fnins}
\BIBentryALTinterwordspacing
T.~Delbruck and M.~Lang, ``Robotic goalie with 3 ms reaction time at 4\% cpu
  load using event-based dynamic vision sensor,'' \emph{Frontiers in
  Neuroscience}, vol.~7, 2013. [Online]. Available:
  \url{http://dx.doi.org/10.3389/fnins.2013.00223}
\BIBentrySTDinterwordspacing

\bibitem{Rebecq2019pami}
\BIBentryALTinterwordspacing
H.~Rebecq, R.~Ranftl, V.~Koltun, and D.~Scaramuzza, ``High speed and high
  dynamic range video with an event camera,'' \emph{{IEEE} Trans. Pattern Anal.
  Mach. Intell. (T-PAMI)}, 2019. [Online]. Available:
  \url{http://rpg.ifi.uzh.ch/docs/TPAMI19_Rebecq.pdf}
\BIBentrySTDinterwordspacing

\bibitem{Gossard2023iros}
\BIBentryALTinterwordspacing
T.~Gossard, J.~Tebbe, A.~Ziegler, and A.~Zell, ``Spindoe: A ball spin
  estimation method for table tennis robot,'' in \emph{2023 IEEE/RSJ
  International Conference on Intelligent Robots and Systems (IROS)}.\hskip 1em
  plus 0.5em minus 0.4em\relax IEEE, Oct. 2023. [Online]. Available:
  \url{http://dx.doi.org/10.1109/IROS55552.2023.10342178}
\BIBentrySTDinterwordspacing

\bibitem{Gossard2024cvprw}
\BIBentryALTinterwordspacing
T.~Gossard, J.~Krismer, A.~Ziegler, J.~Tebbe, and A.~Zell, ``Table tennis ball
  spin estimation with an event camera,'' in \emph{2024 IEEE/CVF Conference on
  Computer Vision and Pattern Recognition Workshops (CVPRW)}.\hskip 1em plus
  0.5em minus 0.4em\relax IEEE, Jun. 2024, p. 3347–3356. [Online]. Available:
  \url{http://dx.doi.org/10.1109/CVPRW63382.2024.00339}
\BIBentrySTDinterwordspacing

\bibitem{Zhang2010tim}
\BIBentryALTinterwordspacing
Z.~Zhang, D.~Xu, and M.~Tan, ``Visual measurement and prediction of ball
  trajectory for table tennis robot,'' \emph{IEEE Transactions on
  Instrumentation and Measurement}, vol.~59, no.~12, p. 3195–3205, Dec. 2010.
  [Online]. Available: \url{http://dx.doi.org/10.1109/TIM.2010.2047128}
\BIBentrySTDinterwordspacing

\end{thebibliography}

\end{document}